\PassOptionsToPackage{twoside=false}{geometry}      

\documentclass[manuscript,screen,nonacm]{acmart}    

\usepackage[ngerman, main=english]{babel}
\usepackage{microtype}
\usepackage[utf8]{inputenc} 
\usepackage[T1]{fontenc} 
\usepackage[yyyymmdd]{datetime} 
\usepackage{mathpazo} 

\usepackage{booktabs}   

\usepackage{graphicx} 
\usepackage{csquotes} 
\usepackage{cleveref} 
\usepackage{xcolor} 
\usepackage{tcolorbox} 
\usepackage{subcaption} 
\usepackage{enumitem} 
\usepackage{placeins} 
\usepackage{hyperref}
\usepackage{todonotes}
\let\xtodo\todo
\renewcommand{\todo}[1]{\xtodo[inline,color=orange!75]{#1}}

\acmConference{Master-Seminar "Advanced Research Methods in Human-Computer Interaction"}{SS 25}{HU Berlin}

\usepackage{listings}
\usepackage{xcolor}
\lstdefinelanguage{json}{
    basicstyle=\ttfamily\small,
    numbers=left,
    numberstyle=\tiny,
    stepnumber=1,
    showstringspaces=false,
    breaklines=true,
    frame=single,
    backgroundcolor=\color{gray!5},
    keywordstyle=\color{blue}\bfseries,
    stringstyle=\color{purple},
    commentstyle=\color{gray}\itshape,
    morestring=[b]",
}

\usepackage{fp}        
\usepackage{siunitx}   

\newcommand\iv[1]{\textsc{#1}}
\newcommand\dataset{\iv{dataset}}
\newcommand\passrate{\iv{PassRate}}
\newcommand\genduration{\iv{GenDuration}}
\newcommand\evalduration{\iv{EvalDuration}}
\newcommand\passduration{\iv{PassDuration}}
\newcommand\rougel{\iv{Rouge-L}}
\newcommand\responselen{\iv{ResponseLen}}

\newcommand\ival[1]{\texttt{‹#1›}}
\newcommand\vanilla{\ival{vanilla}}
\newcommand\tuned{\ival{tuned}}
\newcommand\json{\ival{json}}
\newcommand\markdown{\ival{markdown}}
\newcommand\yaml{\ival{yaml}}

\newcommand\replaceable[1]{\texttt{<#1>}}

\newenvironment{var}[1]{%
  \par\noindent%
  \makebox[3em][l]{}%
  \begin{minipage}[t]{\dimexpr\linewidth-3em}%
  \hspace{-3em}\textbf{#1}\hspace{0.5em}%
}{%
  \end{minipage}\par\vspace{1ex}%
}

\newcommand{\pvalue}[1]{%
  \IfSubStr{#1}{e}{
    \def\pdisplay{< 0.001}
  }{
    \ifdim #1 pt < 0.001 pt
      \def\pdisplay{< 0.001}%
    \else
      \def\pdisplay{= \num[round-precision=3, round-mode=places]{#1}}%
    \fi
  }
  \pdisplay
}
\newcommand{\kruwa}[3]{%
  \FPeval\epsilonsquared{round((#1 - #2)/(820 - #2 - 1),4)}%
  \ifdim \epsilonsquared pt > 0.1379 pt
    \def\effectsize{large}%
  \else
    \ifdim \epsilonsquared pt > 0.0588 pt
      \def\effectsize{medium}%
    \else
      \ifdim \epsilonsquared pt > 0.0099 pt
        \def\effectsize{small}%
      \else
        \def\effectsize{negligible}%
      \fi
    \fi
  \fi
  ($\chi^2$(#2) = \num[round-precision=2, round-mode=places]{#1}, $p \pvalue{#3}$, $\varepsilon^2 = \num[round-precision=3, round-mode=places]{\epsilonsquared}$)%
}
\newcommand{\mawiu}[4]{%
  \edef\r{\fpeval{abs(#3)}}%
  \ifdim \r pt > 0.5 pt
    \def\effectsize{large}%
  \else
    \ifdim \r pt > 0.3 pt
      \def\effectsize{medium}%
    \else
      \ifdim \r pt > 0.1 pt
        \def\effectsize{small}%
      \else
        \def\effectsize{negligible}%
      \fi
    \fi
  \fi
  \ifx &#4&
    \def\unit{}%
  \else
    \ifx#4\%
      \def\unit{\%}%
    \else
      \def\unit{~#4}%
    \fi
  \fi
  ($p \pvalue{#1}$, $\Delta\tilde{x} = \num[round-precision=2, round-mode=places]{#2}$\unit, $r = \num[round-precision=3, round-mode=places]{#3}$)
}

\begin{document}

\title{PromptResponse: Optimizing Prompts for LLM Coding Tasks} 

\author{Erik Thureck}
\email{erik.thureck@hu-berlin.de}
\orcid{0009-0002-1994-5648}
\affiliation{%
  \institution{HU~Berlin}
  \city{Berlin}
  \country{Germany}
}
\author{Robert Kühnen}
\email{robert.kuehnen@hu-berlin.de}
\orcid{0009-0001-6329-0581}
\affiliation{%
  \institution{HU~Berlin}
  \city{Berlin}
  \country{Germany}
}
\author{Tim Jacobowitz}
\email{jacobowt@hu-berlin.de}
\orcid{0009-0009-3350-6095}
\affiliation{%
  \institution{HU~Berlin}
  \city{Berlin}
  \country{Germany}
}





\begin{abstract}
    \textbf{\textsc{Abstract}.} Large language models (LLMs) are increasingly used in research workflows and software development pipelines, yet their output remains sensitive to input prompt variations. This paper presents <<PromptResponse>>, a controlled study examining how formatting and LLM-based tuning of coding task prompts affect the resulting code's performance, efficiency, and stability. Using five semantically identical yet syntactically distinct variants of the HumanEval dataset—baseline, JSON, Markdown, YAML, and an LLM-tuned version—we had GPT-4o solve its coding problems over 8200~executions. Our results show that consistent formatting---especially JSON---improves generation efficiency and syntactic stability, with minor gains in task performance.
    Conversely, the LLM-tuned prompts resulted in significantly degraded task performance without significant improvements in any other dimension.
    These findings suggest that low-effort reformatting alone can yield measurable improvements, while tuning must account for model alignment. We conclude our work with providing a set of practical recommendations informed by our results as well as releasing our dataset variants and evaluation pipeline for future work.
\end{abstract}






\begin{teaserfigure}
  \includegraphics[height=19em]{figures/teaser_blank.png}
\end{teaserfigure}

\vspace*{-3.25cm}
\begin{minipage}[c][3cm][c]{0.97\textwidth}
    \newcommand{\HRule}{\rule{\linewidth}{0.5mm}}
    \newlength{\parindentbak} \setlength{\parindentbak}{\parindent}
    \newlength{\parskipbak} \setlength{\parskipbak}{\parskip}
    \setlength{\parindent}{0pt}
    \setlength{\parskip}{\baselineskip}

    \textsc{%
    \begin{flushright}
        \textls*[68]{\Large Humboldt-Universität zu Berlin}\\
        \normalsize \textls*[45]{%
            Mathematisch-Naturwissenschaftliche Fakultät\\
            Institut für Informatik\\
            Human-Computer Interaction Lab
        }
    \end{flushright}
    }
\end{minipage}

\vspace{3em}

\maketitle

\vspace{-44em}
\begin{figure}[h!]
\centering
\begin{minipage}[t]{0.461\textwidth}
\centering
\begin{lstlisting}[language=json, caption={}, basicstyle=\tiny]
def greet(names):
    """A simbel program which should return a greeting for all names in the input list.
    Note: If no names or or only 'world' is provided, great the stranger or write 'Hello World!'.

    for examble:
    for greet(['Erik']) == 'Hello, Erik!'
    assert greet(['Robert','Tim']) => 'Hello, Robert & Tim!'
    greeting(['world']) # ==> 'Hello World!'
    greet([]) -> 'Hello, stranger!'

    Have fun :)
    """
\end{lstlisting}
\end{minipage}
\hfill
\begin{minipage}[t]{0.461\textwidth}
\centering
\begin{lstlisting}[language=json, caption={}, basicstyle=\tiny]
{
  "function": "greet",
  "signature": "greet(names: list[str]) -> str",
  "description": "A simple program that should return a greeting for all names in the input list.",
  "note": "If no names or only 'world' is provided, greet the stranger or write 'Hello World!'.",
  "examples": [
    {"input": ['Erik'], "output": "Hello, Erik!"},
    {"input": ['Robert','Tim'], "output": "Hello, Robert & Tim!"},
    {"input": ['world'], "output": "Hello World!"},
    {"input": [], "output": "Hello, stranger!"}
  ]
}
\end{lstlisting}
\end{minipage}

\begin{minipage}[t]{0.30\textwidth}
    \raggedleft
    \includegraphics[height=3.5em]{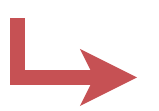}
\end{minipage}
\hfill
\begin{minipage}[t]{0.30\textwidth}
\centering
\vspace{-2.6em}\textbf{LLM-generated Code}

Task Performance, Efficiency,

\& Prompt Stability?
\end{minipage}
\hfill
\begin{minipage}[t]{0.30\textwidth}
    \raggedright
    \reflectbox{\includegraphics[height=3.5em]{figures/arrow.pdf}}
\end{minipage}
\begin{minipage}[t]{\textwidth}
\centering
\vspace{-3.45em}Original \phantom{aaaaaaaaaaaaaaaaaaaaaaaaaaaaaaaaaaaaaaaaaaaaaaaaaaaaaaaaaaaa Original}

\vspace{-1.25em}\phantom{Reformatted aaaaaaaaaaaaaaaaaaaaaaaaaaaaaaaaaaaaaaaaaaaaaaaaaaaaaaaaaaaa} Reformatted
\end{minipage}\vspace{-0.375em}
\caption{A prompt designed to highlight some of the inconsistencies present in the HumanEval dataset, before and after possible reformatting. Which one will lead to better results?}
\label{fig:teaser}
\vspace{-6.71cm}
\end{figure}
\begin{figure}[h!]
  \centering
  \includegraphics[width=15.24cm]{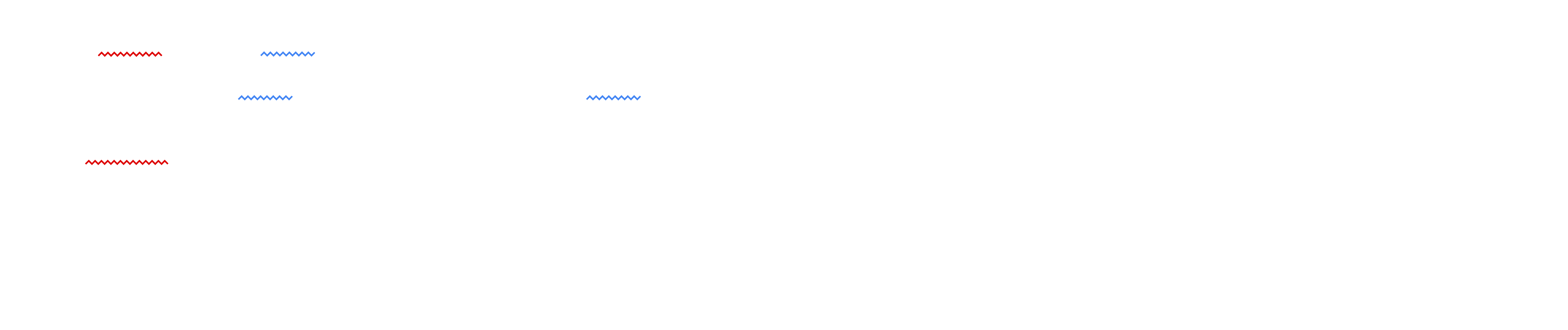}
\end{figure}

\clearpage
\section{Introduction}\label{sec:intro}

With the rapid proliferation of artificial intelligence (AI) and large language models (LLMs) in all walks of life in recent years, they also play an increasingly important role in the field of human-computer interaction (HCI)~\cite{pang2025}. Due to their high accessibility and ease of use via chat interfaces, they have not only caught the interest of civilians but researchers alike, who have begun using them, for example, for text classification, generating synthetic participant data, and even to produce code. However, the integration of LLMs into research workflows is often ad hoc, as researchers tend to formulate prompts on the fly, informally, and without following best practices. Not only does this jeopardize the stability of results, but---especially when not properly documented---heavily undermines the reproducibility of the conducted work.


LLM-generated output is known to be sensitive to variations in input phrasing and format. This raises questions about the scientific validity of LLMs as tools in the field of research. Previous work has proposed best-practice guidelines for prompting LLMs~\cite{Törnberg2024_BestPractices,pang2025}, as well as metrics such as the Prompt Stability Score (PSS)~\cite{Barrie2025_PromptStabilityScoring} or benchmarks like PromptSET~\cite{razavi_benchmarking_2025} and E-Bench~\cite{zhang_e-bench_2024} to quantify this sensitivity. These contributions show that even semantically identical prompts can lead to varying outputs~\cite{atil_non-determinism_2025}, especially under low-control or real-world conditions. LLMs are also increasingly being used for code generation tasks by individual programmers as well as enterprise-level development teams because of their ability to swiftly translate natural language into executable code and support various stages of the software engineering life cycle~\cite{jiang2025_survey}. However, existing research on prompt optimization and stability tends to focus on natural language tasks like question answering or text annotation, leaving prompt optimization techniques for code generation underinvestigated. Moreover, although existing research proposes strategies for tuning prompts to optimize LLM output, automating this process hasn't yet been explored. In particular, it remains to be investigated whether an LLM can autonomously tune prompts by applying a predefined, instructed strategy.







Thus, the goal of this paper is to close these gaps and mature the field of LLM code generation as a whole by investigating different prompt-tuning strategies. Not only could the optimization of prompts increase the performance and efficiency of resulting code, but also prompt stability and output reproducibility, validating the usage of LLMs in research contexts.




The contributions of this paper are threefold: First, we present four syntactically different yet semantically equivalent derivatives of OpenAI's HumanEval dataset---a dataset of 164~coding problems---with improved internal consistency, which we created by parsing the original into the JSON, Markdown, and YAML formats as well as using a different LLM for prompt-wise tuning.
Second, we present the results from a controlled experiment of 8200~GPT-4o API requests investigating the influences of their characteristics on task performance, efficiency, and prompt stability of the LLM-generated code compared to the unaltered original.
Third, based on these findings, we provide implications and guidelines for the deployment of LLMs in coding tasks.




\section{Related Work}\label{sec:related-work}









Recent years have seen a plethora of research on best practices for ethical LLM usage in academic contexts, prompt optimization, stability metrics and how they run afoul of the inherent nondeterminism of LLMs as well as how LLMs can be employed for code generation, for example, in multi-agent systems.
This section provides an overview of the most relevant works in this field.


\subsection{Guidelines for LLM Usage in Research}
In 2025, Pang et al.~\cite{pang2025} systematically reviewed 153~CHI papers on LLMs released between 2020 and 2024. The authors identified where and how LLMs are used within HCI, dividing applications into ten categories, like communication, education, or programming. Five roles LLMs play are outlined, which are system engines, research tools, simulated participants, objects of study, and users' perceptions of LLMs. It is shown that LLMs play a significant role in HCI research, with concerns often raised about validity, reproducibility, and ethical risks. They also advised researchers in the field to release their used prompts and LLM outputs publicly to ensure transparency and reproducibility~\cite{pang2025}.

After LLMs had become increasingly popular for text annotation due to their ease of use, high accuracy, and comparatively low cost, in 2024, Törnberg~\cite{Törnberg2024_BestPractices} proposed a set of guidelines for using LLMs in text annotation to address concerns regarding research quality and integrity.
Aside from recommendations on model selection and the consideration of ethical and legal implications, he discussed prompt engineering, structured prompting, and prompt stability analysis to promote reliable, reproducible, and ethical use whilst mitigating biases and misunderstandings. His guidelines included: simultaneously developing coding instructions and the LLM prompt until sufficient agreement between the LLM and the human encoders is reached, developing a «prompt codebook» describing the LLM prompts and parameters to minimize the disagreement between different coders, and structuring prompts into the sections \textit{context}, \textit{question}, and \textit{constraints}. The latter might include answers having to be formatted in JSON or allowing «I don't know» as an answer. However, his work is only a collection of best practices and included neither an experiment nor a statistical analysis~\cite{Törnberg2024_BestPractices}.

The following year, in 2025, Kosch and Feger~\cite{Kosch2025_PromptHacking} urged against the inflationary application of LLMs in data analysis tasks, as their inherent biases---hailing from their training data---non-deterministic outputs, and hallucinations make them fundamentally unreliable when impartiality and reproducibility would be required.
They further likened <<prompt-hacking>>, the practice of strategically tweaking prompts to elicit desirable LLM outputs, to <<p-hacking>>, where researchers retroactively tune experimental data or test parameters to manufacture statistically significant results, in turn coming to misled conclusions whilst undermining scientific integrity. However, they also noted that unlike p-hacking, prompt-hacking's inherent partiality can invalidate results even under proper usage.
To mitigate these challenges, they recommended preregistering prompts, documenting their modifications, and, in general, forgoing LLMs where possible~\cite{Kosch2025_PromptHacking}.

\subsection{Prompt Stability}\enlargethispage{2em}
In 2025, Barrie et al.~\cite{Barrie2025_PromptStabilityScoring} presented a technique to test for prompt stability in LLMs, analogous to inter- and intra-coder reliability, for text annotation. They outlined an algorithm for generating semantically similar prompts from a baseline prompt---input by a user---and a consistency analysis of the LLM responses within the same and across semantically similar prompts, resulting in the «Prompt Stability Score» (PSS).
Based on the results from their classification of about 3.1~million rows of data and 300~million input tokens among six different datasets and twelve outcomes, they provided best practice recommendations for applied research~\cite{Barrie2025_PromptStabilityScoring}.

The same year, Razavi et al.~\cite{razavi_benchmarking_2025} introduced PromptSET, a benchmark built from semantically equivalent variations of question-answering prompts (TriviaQA and HotpotQA), to study LLM prompt sensitivity. They formalized the task of prompt sensitivity prediction, aiming to predict whether an LLM will answer a prompt variation correctly. Benchmarking several baselines, including self-evaluation, text classification, and query performance prediction, they found that existing methods performed poorly. This underscores the need for more robust tools to assess and improve prompt stability~\cite{razavi_benchmarking_2025}.

The previous year, in 2024, Zhang et al.~\cite{zhang_e-bench_2024} proposed E-Bench, a benchmark evaluating LLM robustness to real-world prompt variations. These included paraphrasing, simplification, colloquial rewording, and typographical noise. Built on AlpacaEval, E-Bench systematically perturbs prompts and measures performance drops across six models, including GPT-4, Llama 2, and Vicuna. Results show that larger models perform better under synonymous changes but struggle with typos. The study underscores the need for human-centered evaluations in low-control settings such as education and casual use~\cite{zhang_e-bench_2024}.

\subsection{Prompt Format}
In 2024, Wang et al.~\cite{wang2024} conducted research on how prompt engineering strategies influence LLM consistency and reliability when answering medical questions. Nine LLMs were tested with various prompt types: IO (Input-Output, direct instructions), 0-COT (Chain-of-Thought, step-by-step reasoning), P-COT (structured reasoning), and ROT (simulated expert discussions). Each question was asked five times for each prompt type. Results showed that \textit{GPT-4$\times$ROT prompting} achieved the highest consistency (62.9\%), while \textit{GPT-3.5$\times$IO~prompting}---with a nearly perfect Fleiss' kappa value of 0.984---demonstrated the best reliability. Their study further provided notable prompt formatting categories and methodologies~\cite{wang2024}.

The same year, Leiter and Eger~\cite{Leiter2024_PrExMe} evaluated more than 720~prompt templates for open-source LLM-based metrics on machine translation and summarization datasets, totaling over 6.6~million evaluations, in what they called «PrExMe» (Prompt Exploration for Metrics). While they discovered scenarios under which prompts were stable, they also found idiosyncratic preferences of some LLMs for certain output formats as well as that seemingly innocuous changes, like shifting the numeric output interval in which the LLM had to answer, could strongly affect the rankings in their evaluation~\cite{Leiter2024_PrExMe}.

Also in 2024, He et al.~\cite{He2024_PromptFormatting} investigated the effect different prompt formats---such as plain text, Markdown, JSON, and YAML---have on the performance of LLMs. While their study revealed prompt formatting to significantly impact the performance of GPT models, they did not uncover any single format to excel universally. Thus, they advised future research to probe diverse prompt formats when testing LLMs in order to paint a more complete picture of their performance. 
However, they did observe a lessened impact of prompt formatting on bigger models like GPT-4, compared to GPT-3.5~\cite{He2024_PromptFormatting}.

\subsection{Non-Determinism of LLMs}\enlargethispage{2em}
In his aforementioned guidelines, Törnberg~\cite{Törnberg2024_BestPractices} also highlighted the importance the choice of the LLM has, as especially closed-source models like ChatGPT are particularly intransparent and changing over time~\cite{Törnberg2024_BestPractices}.
The latter phenomenon was further investigated by Chen et al. in 2024, who found that the accuracy of the same GPT-4 model, for example, at identifying prime vs. composite numbers dropped from 84\% to 51\% within three months from March to June 2023~\cite{Chen2024_ChatGPTChanges}.

In 2025, Atil et al.~\cite{atil_non-determinism_2025} presented a systematic study of non-determinism in LLMs under configurations intended to be deterministic, such as setting the temperature to 0. Using five LLMs across eight tasks from BBH\footnote{Beyond the Imitation Game Benchmark: Hard Subset} and MMLU\footnote{Massive Multitask Language Understanding}, they observed substantial output variability, with up to 15\% accuracy fluctuation across runs and up to 70\% difference between best- and worst-case outcomes. To quantify this, they introduced two agreement metrics: TARr@N\footnote{Total Agreement Rate at N runs} (raw output agreement) and TARa@N\footnote{Total Agreement Rate of parsed answers at N runs} (parsed answer agreement). Their findings highlight a core reproducibility problem in LLM-based research, even under controlled settings. However, they did not report the specific prompts used and explicitly avoided prompt optimization techniques, such as chain-of-thought or instruction tuning. Since prompt design is known to significantly impact both accuracy and consistency~\cite{razavi_benchmarking_2025}, their reported performance likely does not reflect an upper bound and may differ under better-crafted prompts~\cite{atil_non-determinism_2025}.

\subsection{LLMs as Code Generation Tools \& Multi-Agent System «AgentCoder»}
In 2025, Jiang et al.~\cite{jiang2025_survey} conducted a comprehensive survey of 235~papers on code generation LLMs, covering the developments in this field between 2020 and 2024. The authors highlighted how integral LLMs have become to software engineering workflows as well as how prompt design and model alignment are critical factors influencing output quality. This underlines the importance of investigating the influence of prompt structure and format in code generation contexts, given the known LLM sensitivity and strict correctness requirements of coding tasks~\cite{jiang2025_survey}.

In 2023, Huang et al.~\cite{huang_agentcoder_2023} proposed AgentCoder, an LLM-based multi-agent system for code generation, employing multiple LLMs as «agents» that collaborate to perform complex tasks by distributing responsibilities among them. AgentCoder involves three such agents to generate code: a programmer agent generating code using a Chain-of-Thought approach~\cite{wang2024}, a test designer agent independently creating test cases, and a test executor agent that runs these tests and provides feedback to refine the code. To test their approach, they used the HumanEval dataset, which consists of diverse programming challenges, probing problem-solving skills and adaptability. These tasks include canonical solutions and test cases to measure the correctness of generated code. While AgentCoder shows how multi-agent LLM systems can be structured to improve code generation quality, it also highlights the trade-off between performance and computational cost. Each agent in the system contributes to the overall token budget, making multi-agent frameworks inherently resource-intensive. This trade-off was a key motivation for AgentCoder’s own reduction to three agents, compared to more expansive systems~\cite{huang_agentcoder_2023}. However, even this setup remains more complex than a single-agent solution.

\subsection{Summary \& Research Question}

In recent years, LLMs have become an integral part of the lives of civilians and researchers alike.
However, related work has repeatedly raised concerns over the ethicality, reproducibility, and, thus, validity of LLM use in research contexts, especially when done without moderation.
In this regard, prompt stability is a promising metric, which, however, has not yet been thoroughly investigated for LLM code generation---one of the technology's most prominent fields of application.
Whereas many previous approaches to improve LLM-generated code have relied on resource-intensive multi-agent systems, prompt formatting has been shown to significantly affect LLM performance in other contexts whilst also potentially mitigating the structural inconsistencies commonly found even in established datasets such as HumanEval.

Therefore, this work aims to inform a new foundation for efficient and reliable LLM code generation by bridging these realities through investigating the following research question:

\begin{itemize}
    \item \textbf{RQ:} How do \textit{LLM tuning} and \textit{prompt formatting} influence the \textit{task performance}, \textit{efficiency}, and \textit{prompt stability} in HumanEval-style coding tasks?
\end{itemize}










\section{Methodology}\label{sec:methodology}

We conducted a controlled experiment to investigate the influences LLM tuning and the formatting of prompts have on the task performance (pass@1), efficiency, as in the generation and evaluation durations, and prompt stability in LLM coding tasks. In particular, we queried OpenAI's GPT-4o to solve the Python coding problems from the HumanEval~\cite{HumanEval} dataset in five semantically identical variations.

Based on the analysis of previous work in the field of prompt engineering for LLMs, we formulated the following hypotheses to guide our investigation in answering our research question:

\begin{itemize}
    \item \textbf{H\textsubscript{1}:} The input prompts' format or having been LLM-tuned affects the task performance.
    \item \textbf{H\textsubscript{2}:} Consistent prompt formatting and LLM tuning improve the efficiency of LLM code generation.
    \item \textbf{H\textsubscript{3}:} Consistent prompt formatting and LLM tuning improve the efficiency of LLM-generated code.
    \item \textbf{H\textsubscript{4}:} Consistent prompt formatting and LLM tuning lead to higher prompt stability.
\end{itemize}

\subsection{Design}\label{sec:methodology/design}
Following the related work, and due to its premiere usage in both societal and academic contexts, we opted to focus on the predominant ChatGPT in our investigation because we deemed it to yield the most widely applicable and, thus, most useful results. We chose GPT-4o in particular, as it's not only OpenAI's flagship model\footnote{\href{https://help.openai.com/en/articles/7864572-what-is-the-chatgpt-model-selector}{OpenAI Model Comparison} (last accessed July 15, 2025)} but also its most popular due to its generalized skill set and reasonable pricing\footnote{\href{https://platform.openai.com/docs/pricing}{OpenAI API Pricing} (last accessed July 15, 2025)}.

Because it has been shown that even <<deterministic>> LLM settings---like a low temperature---do not result in reliably deterministic and stable output (cf.~\cite{atil_non-determinism_2025}), and due to the focus of this investigation being the formatting of prompts, we decided against using custom settings. Not only do we avoid inserting our own biases into the study design by not choosing specific settings, but we also ensure that our results bear the widest applicability, as most users will never deviate from the default settings at all.

\subsubsection{Independent Variable}\label{sec:methodology/design/ivs}
To gain a comprehensive understanding of how the syntactic properties of the prompts themselves or having been tuned with an LLM beforehand influence task performance, efficiency, and prompt stability, we varied the independent variable \textbf{\dataset{}}---the specific explication of HumanEval's 164 coding problems---between the following five levels:

\begin{enumerate}
    \item The original, unaltered HumanEval dataset in \textbf{\vanilla{}} as the baseline.
\end{enumerate}

\noindent Because ChatGPT---as all LLMs---uses natural language processing (NLP) to tokenize and process the inputs it has been given, their syntactic structure is a promising factor to investigate, with even slight changes that don't cause any change to semantics possibly leading to vastly different outcomes.
We, thus, further parsed the original dataset into the following three human-readable templates:

\begin{enumerate}[start=2]
    \item The HumanEval dataset in \textbf{\json{}} format.
    \item The HumanEval dataset in \textbf{\markdown{}} format.
    \item The HumanEval dataset in \textbf{\yaml{}} format.
\end{enumerate}

\noindent The wording of each prompt's coding task---e.g., the function signature and description---stayed consistent throughout all these variations of \dataset{} so as to not introduce random error.
Fifthly, we investigated:

\begin{enumerate}[start=5]
    \item The HumanEval dataset, but with its docstrings \textbf{\tuned{}} using an LLM by Mistral AI.
\end{enumerate}

\noindent We entered each dataset's 164 prompts separately into virgin ChatGPT windows without any previous conversational context---akin to a between-groups design in a user study---and, therefore, did not have to counterbalance against the effects of order.
Each prompt was queried 10~times to probe for prompt stability, totaling $10 \times 164 = 1640$~executions per \dataset{} and, thus, $5 \times 1640 = 8200$~executions in total.

\subsubsection{Dependent Variables}\label{sec:methodology/design/dvs}
Based on the responses generated by the investigated LLM---OpenAI's GPT-4o---we calculated the following dependent variable as a quantitative measure of task performance and the effective stability over each prompt's 10 executions:

\begin{var}{\passrate{}}
    The ratio of how many of the 10~LLM responses for a \dataset{}'s prompt manage to pass all associated HumanEval tests (pass@1).
\end{var}

\vspace{1em}\noindent We further recorded the following four measures of efficiency during the code generation and evaluation phases for each execution:

\begin{var}{\genduration{}}
    The time it took the LLM to formulate its response---i.e. the solution to the provided prompt's coding problem---from sending the request to the API to receiving the finished response on the host machine.
    Although this measure isn't universally representative due to being dependent on server workload, it might be insightful for time-local comparisons on how efficiently LLMs handle different input formats.
\end{var}

\begin{var}{\evalduration{}}
    The time it took the host machine to evaluate the LLM's response against the HumanEval-provided test cases in Python.
    This is a more classical measure of code efficiency, providing insights on the computational effort and resources required to run LLM-generated code.
\end{var}

\begin{var}{\passduration{}}
    The \evalduration{}, but only for the LLM responses that passed all HumanEval-provided test cases in Python.
    Excluding syntactically and semantically faulty solutions from this measure of code efficiency makes it the more internally valid and real-world applicable version of \evalduration{}.
\end{var}

\begin{var}{\responselen{}}
    The number of characters making up the LLM's solution to a coding problem for each execution.
\end{var}

\vspace{1em}\noindent Lastly, as measures of syntactic prompt stability, we calculated the following dependent variable:

\begin{var}{\rougel{}}
    The average of the pairwise ROUGE-L scores---denoting the longest common subsequence---between all 10~executions of a prompt.
    This value ranges from 0.0 to 1.0 for completely distinct to perfectly matching responses.
\end{var}

\FloatBarrier
\subsection{Datasets}\label{sec:methodology/datasets}
In this section we will go more in depth on the HumanEval dataset~\cite{HumanEval}---which we tested unaltered as \vanilla{}---discuss its various properties and sections, and how we derived the other four levels of \dataset{} from it.

\subsubsection{vanilla}\label{sect:dataset/vanilla}
Each of the 164~\vanilla{} HumanEval prompts specifies a Python coding problem as a function \replaceable{signature} together with a docstring containing a \replaceable{description} (see Listing~\ref{lst:vanilla/min}) as well as the following optional fields: 97.6\% of prompts\footnote{Only prompts \ival{/38}, \ival{/41}, \ival{/50}, and \ival{/83} do not provide examples.} contain \replaceable{examples}---structured \replaceable{input\_i}, expected \replaceable{output\_i}, and optionally an explanatory \replaceable{comment\_i}, for the i\textsuperscript{th} example---following after the \replaceable{description}.
14.0\%---primarily from the first half of prompts---specify \replaceable{imports} that the generated coding solution is expected to utilize, which preface the rest of the prompt.
A further 9.8\% of prompts---although none in the first half---contain an additional \replaceable{note}, which is usually placed between the \replaceable{description} and \replaceable{examples} but in four cases only after the latter\footnote{Of all 16~prompts with a separately specified \replaceable{note}, only \ival{/99}, \ival{/107}, \ival{/120}, and \ival{/160} contain it after the \replaceable{examples} instead of directly following the \replaceable{description}.}.
3.7\% of prompts define a list of special \replaceable{constraints} trailing the prompt.
Four prompts (2.4\%) from the first third\footnote{Prompts \ival{/10}, \ival{/32}, \ival{/38}, and \ival{/50} are prefaced with a helper function to be used in the solution.} include a fully-implemented \replaceable{helper\_function} placed before the main function's requirements, which can be called in the solution.
Only two prompts (\ival{/84} and \ival{/159}) separately explicate its function arguments in \replaceable{variables}---with the components \replaceable{identifier\_i}, \replaceable{type\_i}, and \replaceable{description\_i} for the i\textsuperscript{th} input variable---between the \replaceable{examples} and \replaceable{constraints}.
The theoretically maximal prompt---which doesn't occur in practice, however---can be seen visualized in Listing~\ref{lst:vanilla/max}.

Beyond the described presence or absence of these sections and their partially inconsistent placing, they also are not always introduced with the same textual label, as, for example, <<Constraints:>> is <<Constrain:>> in one case (\ival{/159}), listing multiple constraints.
The \replaceable{examples}, too, have various headers aside from <<Examples:>> like <<Example:>>, <<For example:>>, or even <<for examble[sic!]:>> in one case (\ival{/67}). Furthermore, the \replaceable{examples} themselves aren't formatted consistently: most of them contain the function call followed by one of many different composed arrow symbols and lastly the expected output (as shown in Listings~\ref{lst:vanilla/min} and~\ref{lst:vanilla/max}). However, some prompts also split them over multiple lines---as when calling a function in a terminal---provide them as \phantom{composed lists, or present them fully textually.}
\vspace{-2.33em} 
\begin{figure}[h!]
\centering
\begin{minipage}[b]{0.45\textwidth} 
\onehalfspace
composed lists, or present them fully textually.
Beyond this still, the dataset's prompts contain various spelling issues, partially wrong function signatures\footnotemark{}, and even a <<FIX>> note prefacing prompt \ival{/64}, which is asking for more test cases to be added.

\hspace{0.3em} Overall, the \vanilla{} dataset thus exhibits a high degree of structural inter-prompt variability---does not, however, contain empty sections, making it the lightest level of \dataset{} with a size of only 192~KB.

\endgroup\vspace{1em}

\centering
\begin{lstlisting}[language=json, caption={Minimal \vanilla{} HumanEval prompt}, label=lst:vanilla/min]
def <signature>:
    """
    <description>
    """
\end{lstlisting}
\end{minipage}
\hfill
\begin{minipage}[b]{0.45\textwidth}
\centering
\begin{lstlisting}[language=json, caption={Maximal \vanilla{} HumanEval prompt}, label=lst:vanilla/max, basicstyle=\tiny]
<imports>


<helper_function>


def <signature>:
    """
    <description>

    Note: <note>
    
    Examples:
    * <function>(<input_1>) -> <output_1> # <comment_1>
    * <function>(...) -> ...              # ...
    * <function>(<input_n>) -> <output_n> # <comment_n>
    
    Variables:
    @<identifier_1> : <type_1>
        <description_1>
    @... : ...
        ...
    @<identifier_m> : <type_m>
        <description_m>
    
    Constraints:
    * <constraint_1>
    * ...
    * <constraint_k>
    """
\end{lstlisting}
\end{minipage}
\end{figure}
\footnotetext{Prompts \ival{/81} and \ival{/149} give examples of a different signature than previously defined.}

\FloatBarrier
\begin{figure}[h!]
\centering
\begin{minipage}[b]{0.45\textwidth}
\onehalfspace
\subsubsection{json}\label{sect:dataset/json}
We decided to keep our parsed levels of \dataset{}---including \json{}---with the highest possible structural inter-prompt similarity to establish equal grounds between the executions of all prompts. Sections that did not appear in \vanilla{} and thus couldn't be parsed without sentient intervention or oversight remain empty, as can be seen in Listings~\ref{lst:json/min} and~\ref{lst:json/max}. The textual contents of all sections---except their introductory labels---were parsed as is, without grammar mistakes being corrected.

\hspace{0.8em} Overall, because all prompts include all section headers---even when empty---the \json{}-formatted level of \dataset{} takes up 233~KB of storage, making it the biggest one in the line-up.

\endgroup\vspace{1em}

\centering
\begin{lstlisting}[language=json, caption={Minimal \json{} HumanEval prompt}, label=lst:json/min]
{
  "function": "<function>",
  "imports": "",
  "helper_function": "",
  "signature": "<signature>",
  "description": "<description>",
  "note": "",
  "examples": [],
  "variables": "",
  "constraints": ""
}
\end{lstlisting}
\end{minipage}
\hfill
\begin{minipage}[b]{0.45\textwidth}
\centering
\begin{lstlisting}[language=json, caption={Maximal \json{} HumanEval prompt}, label=lst:json/max, basicstyle=\footnotesize]
{
  "function": "<function>",
  "imports": "<imports>",
  "helper_function": "<helper_function>",
  "signature": "<signature>",
  "description": "<description>",
  "note": "<note>",
  "examples": [
    {
      "input": "<input_1>",
      "output": "<output_1>",
      "comment": "<comment_1>"
    },
    {...},
    {
      "input": "<input_n>",
      "output": "<output_n>",
      "comment": "<comment_n>"
    }
  ],
  "variables": [
    {
      "identifier": "<identifier_1>",
      "type": "<type_1>",
      "description": "<description_1>"
    },
    {...},
    {
      "identifier": "<identifier_m>",
      "type": "<type_m>",
      "description": "<description_m>"
    }
  ],
  "constraints": [
    "<constraint_1>",
    "...",
    "<constraint_k>"
  ]
}
\end{lstlisting}
\end{minipage}
\end{figure}

\FloatBarrier
\subsubsection{markdown}\label{sect:dataset/markdown}
\begin{figure}[h!]
\centering
\begin{minipage}[t]{0.45\textwidth}
\centering
\begin{lstlisting}[language=json, caption={Minimal \markdown{} HumanEval prompt}, label=lst:markdown/min, basicstyle=\scriptsize]
## Function: `<function>`

**Imports**

_None_

**Helper Function**

_None_

**Signature**

`<signature>`

**Description**

<description>

**Note**

_None_

**Examples**

_None_

**Variables**

_Not specified_

**Constraints**

_None_
\end{lstlisting}
\end{minipage}
\hfill
\begin{minipage}[t]{0.45\textwidth}
\centering
\begin{lstlisting}[language=json, caption={Maximal \markdown{} HumanEval prompt}, label=lst:markdown/max, basicstyle=\tiny]
## Function: `<function>`

**Imports**

<imports>

**Helper Function**

<helper_function>

**Signature**

`<signature>`

**Description**

<description>

**Note**

<note>

**Examples**
| Input | Output | Comment |
|-------|--------|---------|
| `<input_1>` | `<output_1>` | <comment_1> |
| `...` | `...` | ... |
| `<input_n>` | `<output_n>` | <comment_n> |

**Variables**
| Identifier | Type | Description |
|------------|------|-------------|
| `<identifier_1>` | <type_1> | <description_1> |
| `...` | ... | ... |
| `<identifier_m>` | <type_m> | <description_m> |

**Constraints**
- `<constraint_1>`
- `...`
- `<constraint_k>`
\end{lstlisting}
\end{minipage}
\end{figure}
As described above for \json{}, the \markdown{} dataset includes all section headers---even when empty---and retains grammar mistakes from the original. However, in this case we denoted not applicable ones as <<\textit{None}>> and \replaceable{variables} as <<\textit{Not specified}>> if not provided, as can be seen in Listings~\ref{lst:markdown/min} and~\ref{lst:markdown/max}.
In general, we used different header formats and tables as well as underscores and backticks to highlight specific values via italics or code boxes, as would be done in real-world applications, when the markdown is rendered for human interpretation.

Overall, the \markdown{}-formatted level of \dataset{} comes out as the second biggest at a size of 218~KB.

\FloatBarrier
\begin{figure}[h!]
\centering
\begin{minipage}[b]{0.45\textwidth}
\onehalfspace
\subsubsection{yaml}\label{sect:dataset/yaml}
The \yaml{} dataset includes the exact same contents as \json{}, likewise leaving empty sections blank instead of filling them in with human-readable placeholder values like we implemented for \markdown{}, as can be seen in the Listings~\ref{lst:yaml/min} and~\ref{lst:yaml/max}.

\hspace{0.8em} Overall, with a size of only 209~KB, \yaml{} is the second smallest level of \dataset{} and shows the most conservative size increase of all three reformatted ones, even though it, too, always includes all headers.

\endgroup\vspace{1em}

\centering
\begin{lstlisting}[language=json, caption={Minimal \yaml{} HumanEval prompt}, label=lst:yaml/min, basicstyle=\footnotesize]
function: <function>
imports: ''
helper_function: ''
signature: '<signature>'
description: <description>
note: ''
examples: []
variables: ''
constraints: ''
\end{lstlisting}
\end{minipage}
\hfill
\begin{minipage}[b]{0.45\textwidth}
\centering
\begin{lstlisting}[language=json, caption={Maximal \yaml{} HumanEval prompt}, label=lst:yaml/max, basicstyle=\footnotesize]
function: <function>
imports: <imports>
helper_function: "<helper_function>"
signature: <signature>
description: <description>
note: <note>
examples:
- input: <input_1>
  output: '<output_1>'
  comment: '<comment_1>'
- ...
- input: <input_n>
  output: '<output_n>'
  comment: '<output_n>'
variables:
- identifier: <identifier_1>
  type: <type_1>
  description: <description_1>
- ...
- identifier: <identifier_m>
  type: <type_m>
  description: <description_m>
constraints:
- <constraint_1>
- ...
- <constraint_k>
\end{lstlisting}
\end{minipage}
\end{figure}

\FloatBarrier
\begin{figure}[h!]
\centering
\begin{minipage}[t]{0.45\textwidth}
\begin{lstlisting}[language=json, caption={Minimal \tuned{} HumanEval prompt}, label=lst:tuned/min]
def <signature>:
    """
    <description*>
    """
\end{lstlisting}

\vspace{1em}
\onehalfspace
\subsubsection{tuned}\label{sect:dataset/tuned}
For \tuned{} we altered the docstrings of the unaltered \vanilla{} prompts by prompting Mistral AI's\footnotemark{} \href{https://huggingface.co/mistralai/Mistral-7B-Instruct-v0.2}{<<mistralai/Mistral-7B-Instruct-v0.2>> model} with the instructions listed in Table~\ref{tab:tuning_prompt}.
The model was used to rewrite all docstrings in the dataset, replacing the originals, as can be seen in Listings~\ref{lst:tuned/min} and~\ref{lst:tuned/max}. During this process, we also standardized the formatting to address inconsistencies present in the original data.
We selected Mistral-7B-Instruct-v0.2 due to \phantom{it being an instruction-tuned large} 

\endgroup

\end{minipage}
\hfill
\begin{minipage}[t]{0.45\textwidth}
\centering
\begin{lstlisting}[language=json, caption={Maximal \tuned{} HumanEval prompt}, label=lst:tuned/max, basicstyle=\tiny]
<imports>


<helper_function>


def <signature>:
    """
    <description*>

    Note: <note*>
    
    Examples:
    * <function>(<input_1>) -> <output_1> # <comment_1*>
    * <function>(...) -> ...              # ...
    * <function>(<input_n>) -> <output_n> # <comment_n*>
    
    Variables:
    @<identifier_1> : <type_1>
        <description_1*>
    @... : ...
        ...
    @<identifier_m> : <type_m>
        <description_m*>
    
    Constraints:
    * <constraint_1*>
    * ...
    * <constraint_k*>
    """
\end{lstlisting}
\end{minipage}
\vspace{-3.15em}
\end{figure}
\footnotetext{\href{https://mistral.ai/}{Mistral AI Homepage} (last accessed July 12, 2025)}

\noindent it being an instruction-tuned large language model, which is freely available and could be efficiently deployed on HU Berlin's \texttt{gruenau8.informatik.hu-berlin.de} server. For this task we used the default temperature of the model.

\begin{table}[h!]
    \centering
    \caption{The messages each MistralAI API request for creating the \tuned{} \dataset{} consisted of.}
    \begin{tabular}{ccp{10cm}}
        \toprule
        \textbf{\#} &\textbf{Role} & \textbf{Content} \\
        \midrule\midrule
        0 & system & You are an expert prompt rewriter. Your task is to improve the clarity and helpfulness of docstrings for code generation. Only rewrite the docstring (the text inside triple double quotes: \textbackslash"\textbackslash"\textbackslash" ... \textbackslash"\textbackslash"\textbackslash"). Do not modify the function signature or write any implementation code. \\
        \midrule
        1 & user & \replaceable{prompt} \\
        \bottomrule
    \end{tabular}
    \label{tab:tuning_prompt}
\end{table}

Overall, the LLM-\tuned{} level of \dataset{}---with its reformulated and extended docstrings---takes up 212~KB of storage, making it the third largest after \json{} and \markdown{}.

\FloatBarrier
\subsection{Procedure}\label{sec:methodology/procedure}
Firstly, we acquired the \vanilla{} instance of the HumanEval dataset from Hugging Face\footnote{\href{https://huggingface.co/datasets/openai/openai_humaneval}{OpenAI/HumanEval Dataset on Hugging Face} (last accessed July 12, 2025)}, parsed it in order to adapt it to the \json{}, \markdown{}, and \yaml{} formats as well as used Mistral AI to generate the \tuned{} variant (see Section~\ref{sec:methodology/datasets}), resulting in the five levels of our independent variable \dataset{}.

\begin{table}[h!]
    \centering
    \caption{The messages each API request consisted of, with a prefix for orientating the LLM before any given \replaceable{prompt}.}
    \begin{tabular}{ccp{10cm}}
        \toprule
        \textbf{\#} &\textbf{Role} & \textbf{Content} \\
        \midrule\midrule
        0 & system & You are a Python programming expert. \\
        \midrule
        1 & user & Please solve the following problem. Output only the function with the signature as specified in the prompt and all necessary imports. Do not include additional texts or comments.\newline\newline\replaceable{prompt} \\
        \bottomrule
    \end{tabular}
    \label{tab:prompt_prefix}
\end{table}

Then, one \dataset{} after the other, we repeated the following procedure using a Python script we had written:
Given one prompt of a \dataset{}, we sent 10 individual ChatGPT API requests, each into a newly-created chat completion window of OpenAI's GPT-4o\footnote{The experiment was conducted using the <<\texttt{gpt-4o-2024-08-06}>> release of OpenAI's GPT-4o.} without any previous context or interactions. Each prompt was always prefaced as depicted in Table~\ref{tab:prompt_prefix} to orientate the model. Once all 10~executions of a prompt---and all 164 prompts of a \dataset{}---had been completed, we continued with the next one until all $5 \times 164 \times 10 = 8200$~executions had been processed.
We disabled response streaming in our API requests so the LLM would only respond with its final solution to the coding problem at hand, without sharing in-progress versions. In case it provided multiple answers to the problem, we always selected the first one.
All responses, together with their recorded \genduration{}, were logged as soon as they had been received on the host device to prevent data loss.

The experiment was conducted on the evening of July 10, 2025, with all executions having been performed in sequential order without multiprocessing, and took about 205~minutes in total.

Afterwards, we used a second Python script to individually read all generated responses, which consisted only of the completed function and potential imports. Once it had cleaned them\footnote{I.e. strip them of the «\texttt{\textasciigrave\textasciigrave\textasciigrave python}» and «\texttt{\textasciigrave\textasciigrave\textasciigrave}» pre- and suffixes the LLM sometimes included.}, it appended them to their respective helper functions from the \vanilla{} dataset, if applicable, and tried to dynamically evaluate them against its test cases using Python's \texttt{exec()} function.
It logged success or failure together with the generated code's \responselen{} and its \evalduration{}. Based on this, it further calculated the \passrate{}, \passduration{}, and \rougel{} score over each prompt's 10~executions (see Section~\ref{sec:methodology/design/dvs}) for later analysis.

By following this procedure, we ensured absolute neutrality of the LLM beyond the prompt at hand and possible biases from its training data, without introducing any external confounding variables.

\subsection{Analysis}
We tested all dependent variables for normality using the Shapiro--Wilk test, which returned highly significant results in all cases.
Consequently, we employed the following non-parametric tests as described below:

We analyzed the recorded data---aggregated across the 10~executions for each \dataset{}'s 164~prompts---using Kruskal--Wallis tests to unveil significant main effects across the five HumanEval variants.
Where these highlighted significant differences, pairwise Mann--Whitney U tests with Bonferroni correction were conducted as post hoc analysis.
We report the epsilon-squared $\varepsilon^2$ as an estimate of the effect size for the Kruskal--Wallis tests, classified based on Cohen's suggestions as small (> 0.0099), medium (> 0.0588), or large (> 0.1379) (cf.~\cite[pp.~285–287]{StatisticalPowerAnalysis1988}).
For the post hoc Mann-Whitney U tests, we further report the rank-biserial correlation $r$ as a measure of the effect size, classified as small (> 0.10), medium (> 0.30), or large (> 0.50)---likewise as per Cohen's suggestions (cf.~\cite[pp.~79–81]{StatisticalPowerAnalysis1988}).

\section{Results}\label{sec:results}
In this section, we report the results from the controlled experiment as described in the previous section.
For each dependent variable, we will provide a summary of the main results along with the findings from the statistical significance analyses conducted in R~\cite{R2024}.

For the \iv{Gen-}, \iv{Eval-}, and \passduration{}s as well as the \responselen{}, where aggregating over all 8200~data points compared to aggregating over the intra-prompt averages---i.e. 820~data points, one for each prompt's 10~executions---makes a difference, we will report the partially aggregated standard deviation $s_{intra}$ alongside $s$, the global one.

\subsection{Pass Rate}
\begin{figure}
  \centering
  \includegraphics[width=\textwidth]{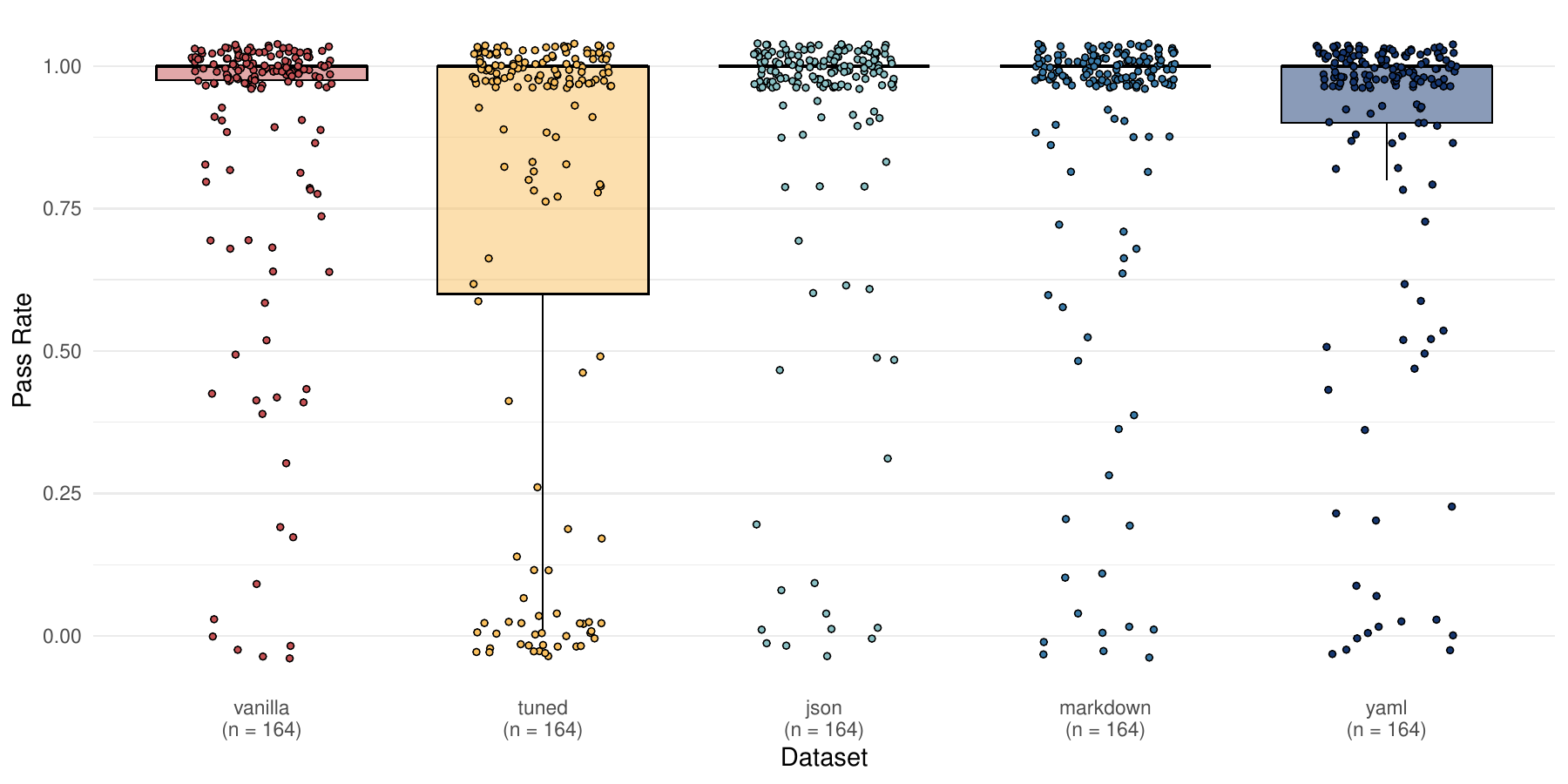}
  \caption{The \passrate{} across all five levels of \dataset{}, with each dot representing the aggregate result of a prompt's 10~executions.}
  \label{fig:PassRate}
\end{figure}
As a measure of task performance and semantic stability, we calculated the \passrate{} for each prompt as the proportion of its 10~executions that successfully passed all test cases from the HumanEval dataset.
Over all 820~prompts from the five levels of \dataset{}, we found 74\% with a perfect \passrate{} of $\frac{10}{10}$, followed by 7.4\% $\frac{0}{10}$, 5.9\% $\frac{9}{10}$, and 3.4\% $\frac{8}{10}$. The remaining ratios---except the rarest, $\frac{3}{10}$ with only 0.5\%---ranged between 1\% and 2\%.
We found the highest perfect \passrate{} rates for \json{} (79.9\%) and \markdown{} (78.7\%), followed by \vanilla{} (75\%) and \yaml{} (73.2\%), with \tuned{} in last (63.4\%), as can be seen in Figure~\ref{fig:PassRate}. Similarly, \vanilla{} had the least total failures (3.7\%), followed by \json{} and \markdown{} (both 4.9\%), \yaml{} (5.5\%), and lastly \tuned{} (18.3\%).
The total average \passrate{}s of \json{} (0.901), \markdown{} (0.890), \vanilla{} (0.886), and \yaml{} (0.873) were similarly high, with \tuned{} (0.748) significantly behind.

The analysis revealed a significant \kruwa{19.1772350857927}{4}{0.0007253633} main effect of \dataset{} with a small effect size.
Post hoc tests confirmed significantly lower pass rates for \tuned{} compared to \json{} \mawiu{0.002122}{0}{-0.162}{}, \markdown{} \mawiu{0.006367}{0}{-0.151}{}, and \vanilla{} \mawiu{0.04545}{0}{-0.128}{} with small effect sizes each.

\subsection{Generation Duration}
\begin{figure}
  \centering
  \includegraphics[width=\textwidth]{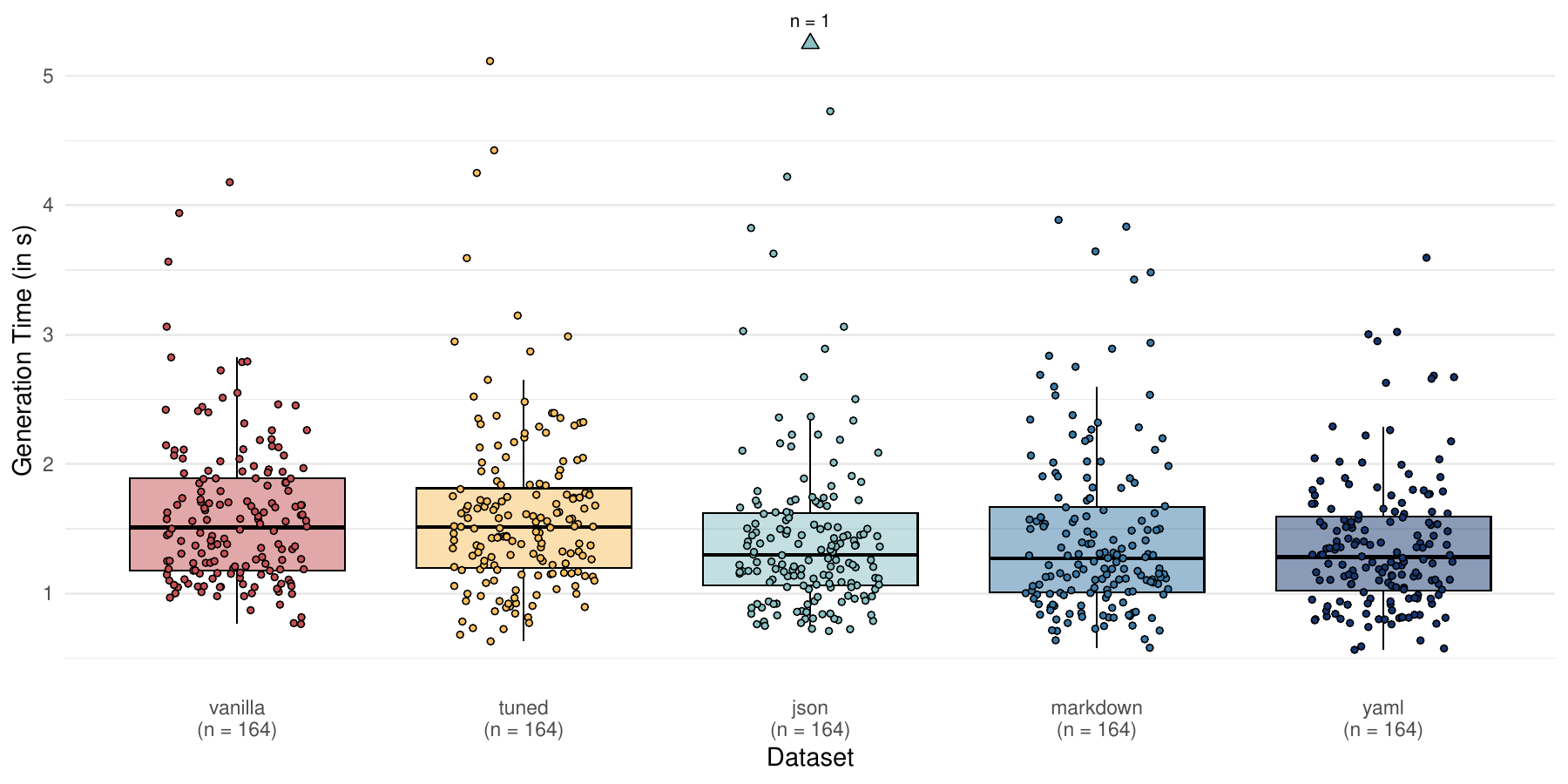}
  \caption{The \genduration{} across all five levels of \dataset{}, aggregated over the 10~executions per prompt.}
  \label{fig:GenDuration}
\end{figure}
As a comparative measure of efficiency, we recorded \genduration{} as the time it took the LLM to formulate its solution to the coding problem specified in a prompt.
We found values ranging from 0.41~s (\ival{yaml/120}) to 20.95~s (\ival{tuned/7}), with one extreme outlier at 95.72~s (\ival{json/80}), as can be seen in Figure~\ref{fig:GenDuration}.
On average, prompts in the \yaml{} format were generated the fastest ($\bar{x} = 1.37$~s, $s_{intra} = 0.36$~s, $s = 0.68$~s), followed closely by \markdown{} ($\bar{x} = 1.44$~s, $s_{intra} = 0.46$~s, $s = 0.91$~s) and \json{} ($\bar{x} = 1.49$~s, $s_{intra} = 0.65$~s, $s = 2.51$~s), with \vanilla{} ($\bar{x} = 1.61$~s, $s_{intra} = 0.55$~s, $s = 0.85$~s) and \tuned{} ($\bar{x} = 1.61$~s, $s_{intra} = 0.57$~s, $s = 1.04$~s) taking significantly longer.

The analysis revealed a significant \kruwa{35.0462004592418}{4}{0.0000004544938} main effect of \dataset{} with a small effect size.
Post hoc tests confirmed significantly higher generation times for \vanilla{} compared to \json{} \mawiu{0.002371}{0.2103}{0.203}{s}, \markdown{} \mawiu{0.002019}{0.2360}{0.205}{s}, and \yaml{} \mawiu{0.0001638}{0.2282}{0.238}{s} with small effect sizes each.
Additionally, \tuned{}, too, led to significantly higher generation times compared to \json{} \mawiu{0.003938}{0.2136}{0.196}{s}, \markdown{} \mawiu{0.00486}{0.2393}{0.193}{s}, and \yaml{} \mawiu{0.0005311}{0.2314}{0.223}{s} with similarly small effect sizes.

\subsection{Evaluation Duration}\label{sect:results/EvalDuration}
\begin{figure}
  \centering
  \includegraphics[width=\textwidth]{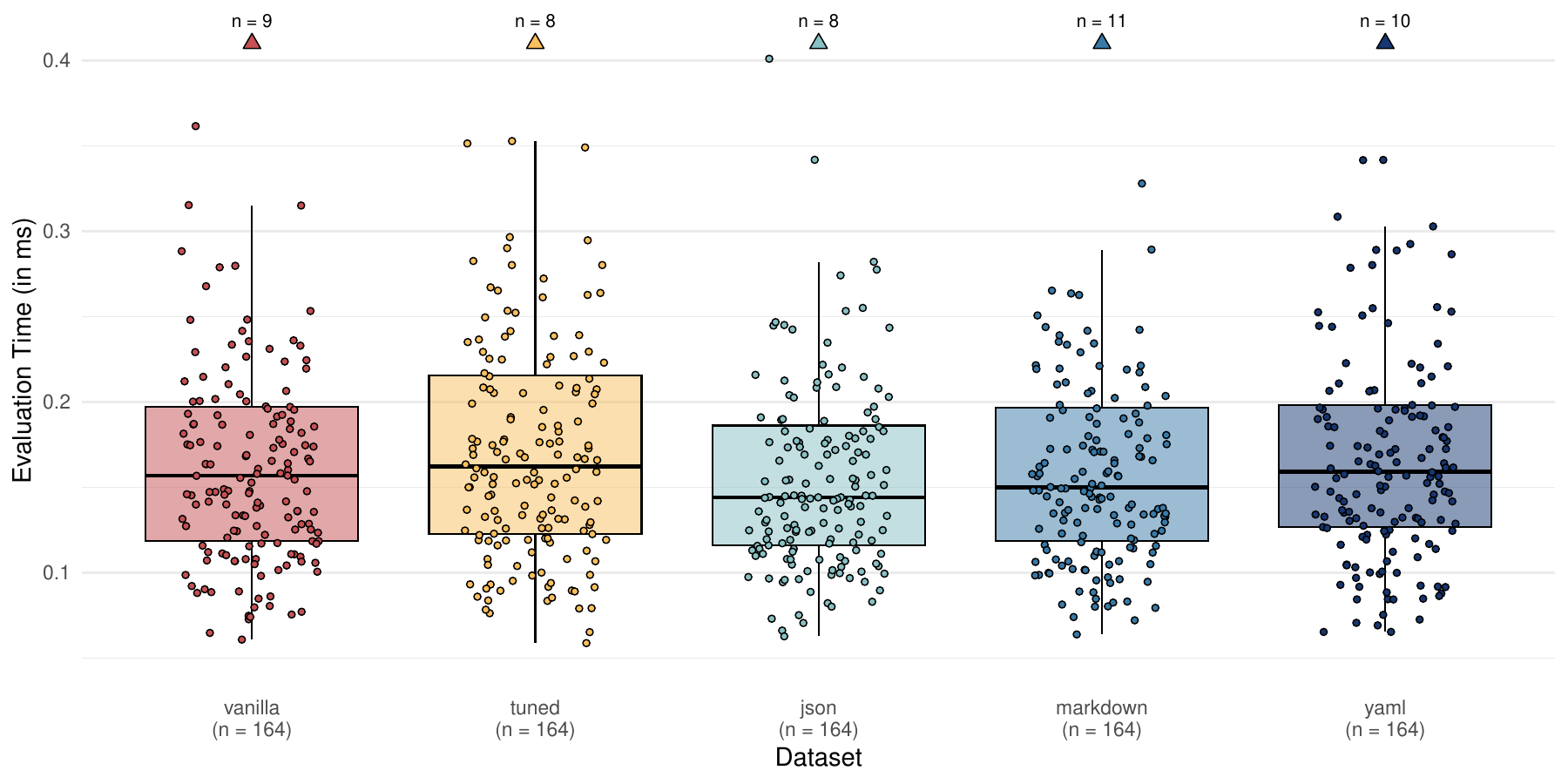}
  \caption{The \evalduration{} across all five levels of \dataset{}, aggregated over the 10~executions per prompt.}
  \label{fig:EvalDuration}
\end{figure}
As a measure of code efficiency, we further recorded \evalduration{} as the time it took to run the LLM-generated Python code against the HumanEval-provided test cases, which we aggregated over all 10~executions of a prompt.
We found values ranging from 48.3~\textmu s (\ival{json/45}) to 0.16~s (\ival{yaml/113}) for all prompts except \ival{/129}, which took between 1.8~s and 17.43~s for 24 out of 40~executions for all formats except \tuned{}\footnote{\ival{HumanEval/129}---or \texttt{minPath(grid, k)}---has a more than 200-word-long description and a \passrate{} of 0.54, with its correct solutions tending to take longer to evaluate. The formats' individual \evalduration{}s were: \vanilla{} (0.52~ms--17.43~s, $\frac{4}{10}$), \json{} (1.91--6.66~s, $\frac{10}{10}$), \markdown{} (0.56~ms--9.93~s, $\frac{7}{10}$), \yaml{} (0.52~ms--7.37~s, $\frac{5}{10}$)---\tuned{} (0.46--3.42~ms, $\frac{1}{10}$) passed only in its longest execution.}.
We further found one outlier in \ival{markdown/91}, which implemented exponential growth with list operations, terminating in a memory error after 31~min 9~s. The distribution without the most extreme of outliers can be seen in Figure~\ref{fig:EvalDuration}.
On average, prompts in the \tuned{} format were evaluated the fastest ($\bar{x} = 0.29$~ms, $s_{intra} = 0.17$~ms, $s = 1.36$~ms), followed at a distance by \yaml{} ($\bar{x} = 14.62$~ms, $s_{intra} = 17.95$~ms, $s = 287.81$~ms) as well as almost equal \json{} ($\bar{x} = 28.55$~ms, $s_{intra} = 10.22$~ms, $s = 383.26$~ms) and \vanilla{} ($\bar{x} = 28.96$~ms, $s_{intra} = 38.28$~ms, $s = 609.73$~ms), with \markdown{} ($\bar{x} = 1.16$~s, $s_{intra} = 3.44$~s, $s = 46.17$~s) in last. Excluding the singular outlying execution of \ival{markdown/91}, \markdown{} would have been the third fastest ($\bar{x} = 15.55$~ms, $s_{intra} = 21.56$~ms, $s = 335.82$~ms).

The analysis revealed no significant differences \kruwa{5.04075265347593}{4}{0.2831415} in \evalduration{} across \dataset{} with a negligible effect size.

\subsection{Pass Duration}
\begin{figure}
  \centering
  \includegraphics[width=\textwidth]{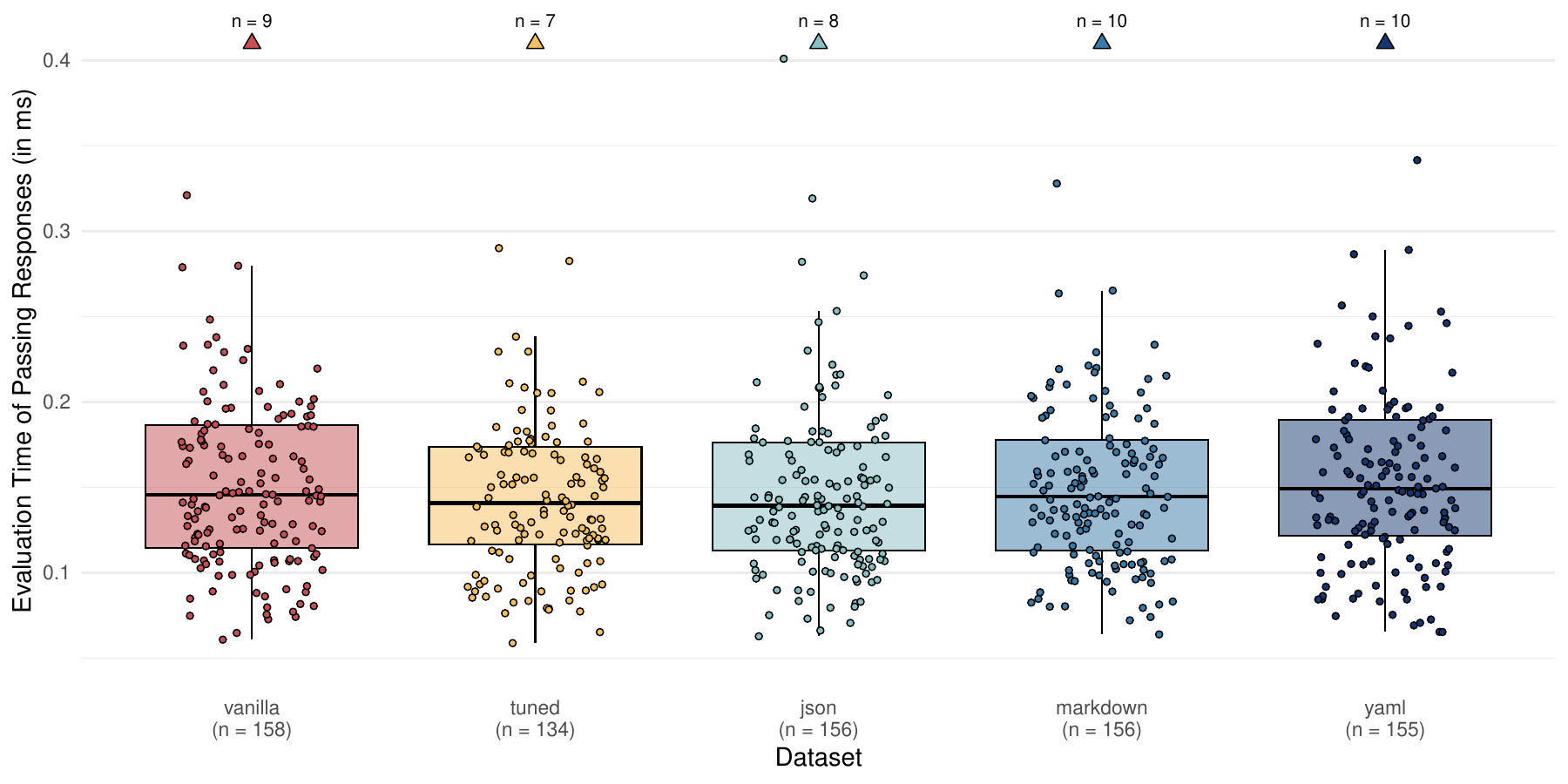}
  \caption{The \passduration{} across all five levels of \dataset{}, aggregated over the up to 10~passing executions per prompt.}
  \label{fig:PassDuration}
\end{figure}
As a second measure of code efficiency, we further recorded \passduration{} as the time it took the LLM-generated Python code to pass all HumanEval-provided test cases---with non-passing executions being excluded---which we aggregated over all up to 10~passing executions of a prompt.
We found values ranging between the same values as \evalduration{}---from 48.3~\textmu s (\ival{json/45}) to 0.16~s (\ival{yaml/113}) for all prompts excluding \ival{/129}, as described above---which can be seen in Figure~\ref{fig:PassDuration}.
Only the upper bound, in the form of the one extreme outlying execution of \ival{markdown/91}, has been reduced, as it did not pass its test cases.
On average, prompts in the \tuned{} format were evaluated the fastest ($\bar{x} = 0.35$~ms, $s_{intra} = 0.18$~ms, $s = 1.58$~ms), followed by \markdown{} ($\bar{x} = 23.32$~ms, $s_{intra} = 24.03$~ms, $s = 355.78$~ms), \json{} ($\bar{x} = 30.01$~ms, $s_{intra} = 10.73$~ms, $s = 403.47$~ms), and \vanilla{} ($\bar{x} = 30.47$~ms, $s_{intra} = 15.75$~ms, $s = 307.96$~ms), with \vanilla{} ($\bar{x} = 74.68$~ms, $s_{intra} = 24.49$~ms, $s = 647.72$~ms) distantly in last.

The analysis revealed no significant differences \kruwa{3.94629794653805}{4}{0.4133224} in \passduration{} across \dataset{} with a negligible effect size.

\subsection{Response Length}
\begin{figure}
  \centering
  \includegraphics[width=\textwidth]{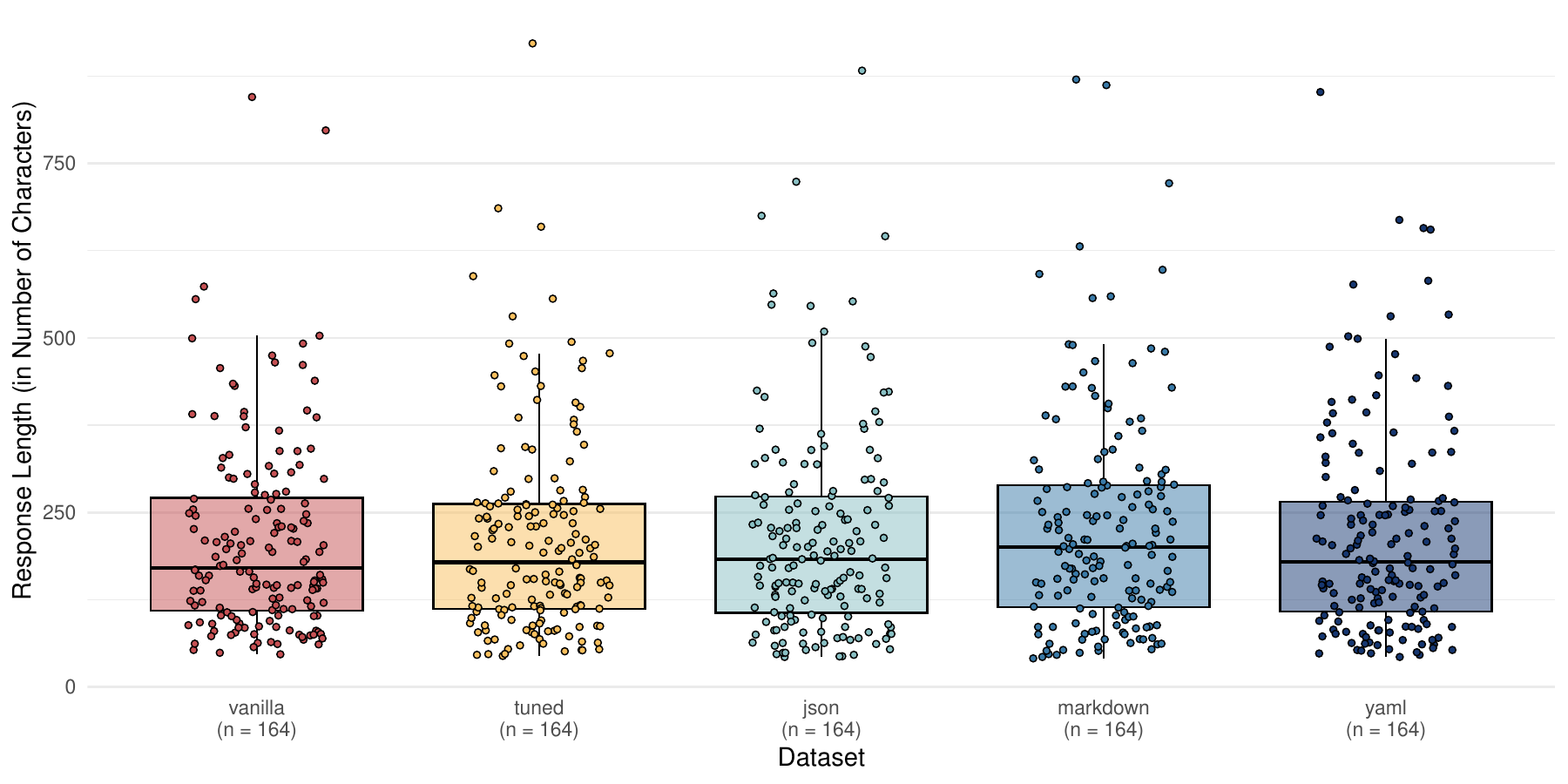}
  \caption{The \responselen{} across all five levels of \dataset{}, aggregated over the 10~executions per prompt.}
  \label{fig:ResponseLen}
\end{figure}
As a further measure of code efficiency, we measured the \responselen{} of each LLM-generated code fragment in its number of characters, which we, too, aggregated over all 10~executions of a prompt.
We found values ranging from 41~chars 25~times for \ival{/53}~($\frac{10}{10}$~\markdown{}, $\frac{8}{10}$~\vanilla{}, $\frac{6}{10}$~\json{}, and $\frac{1}{10}$~\tuned{}, all of which passed) to 1321~chars (\ival{tuned/129}, which did not pass), as can be seen in Figure~\ref{fig:ResponseLen}.
On average, \vanilla{} had the shortest responses ($\bar{x} = 209.4$, $s_{intra} = 44.4$, $s = 152.6$) and \markdown{} the longest ($\bar{x} = 228.1$, $s_{intra} = 33.4$, $s = 162.1$). The responses of \tuned{} ($\bar{x} = 212.0$, $s_{intra} = 37.8$, $s = 155.2$), \json{} ($\bar{x} = 213.5$, $s_{intra} = 27.9$, $s = 152.1$), and \yaml{} ($\bar{x} = 214.9$, $s_{intra} = 28.5$, $s = 152.4$) were similarly short on average.

The analysis revealed no significant differences \kruwa{1.37200562632649}{4}{0.8490457} in \responselen{} across \dataset{} with a negligible effect size.

\subsection{ROUGE-L Scores}
\begin{figure}
  \centering
  \includegraphics[width=\textwidth]{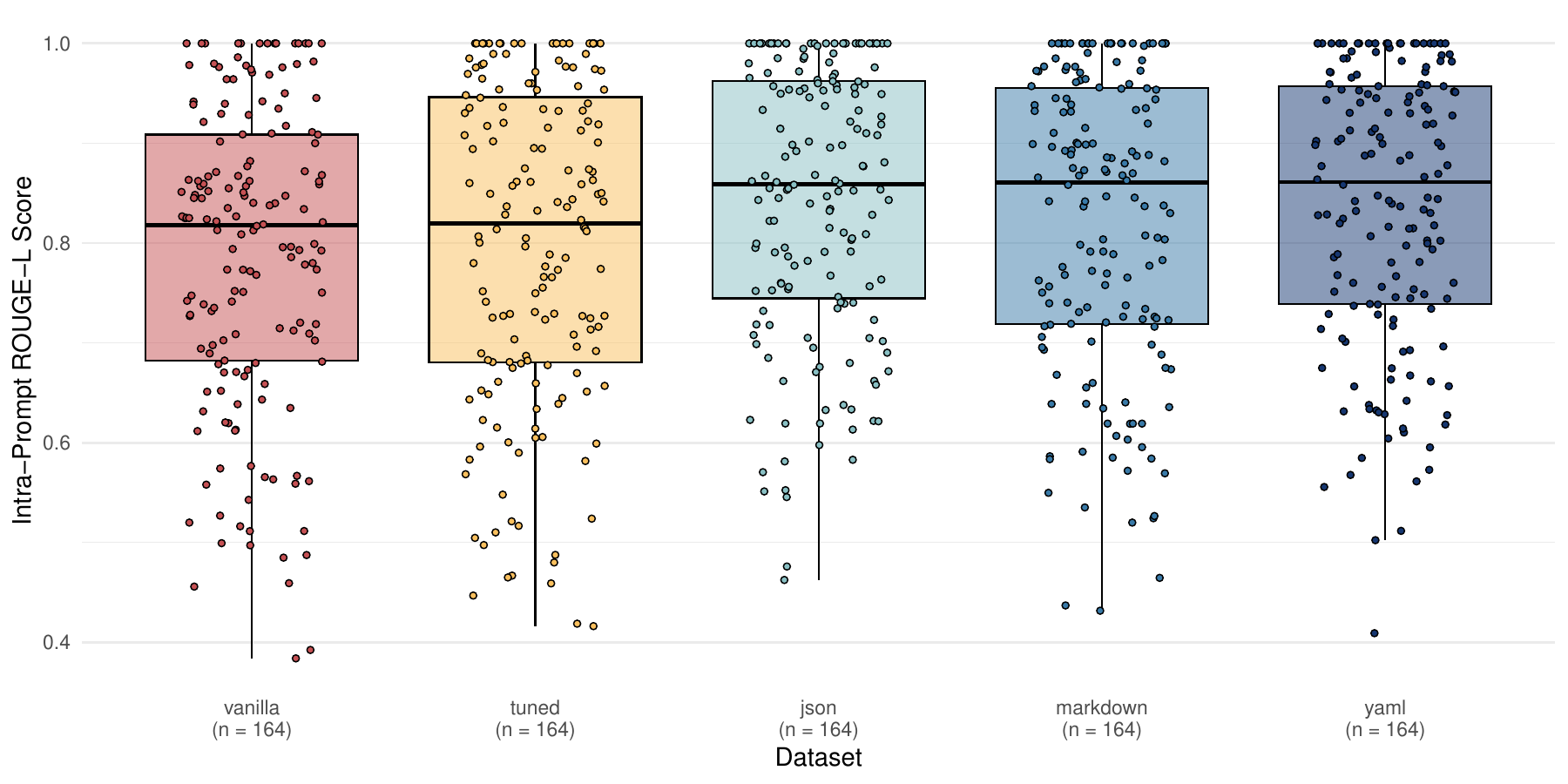}
  \caption{The averages of the pairwise \rougel{} scores across all five levels of \dataset{}, with each dot representing a prompt's 10~executions.}
  \label{fig:RougeL}
\end{figure}
As a measure of syntactic prompt stability, we calculated the average of the pairwise \rougel{} scores between the responses from all 10~executions of a prompt.
We found values ranging from 0.384 (\ival{vanilla/93}) to 1.0 102~times, i.e. for 12.4\% of a prompt's 10~executions (26$\times$\json{} (15.9\%), 22$\times$\markdown{} (13.4\%), 21$\times$\yaml{} (12.8\%), 19$\times$\tuned{} (11.6\%), and 14$\times$\vanilla{} (8.5\%)), as can be seen in Figure~\ref{fig:RougeL}.
On average, reformatted prompts resulted in higher \rougel{} scores, with \json{} ($\bar{x} = 0.842$, $s_{intra} = 0.136$) in first, followed by \markdown{} ($\bar{x} = 0.837$, $s_{intra} = 0.138$) and \yaml{} ($\bar{x} = 0.821$, $s_{intra} = 0.148$). The average scores of \tuned{} ($\bar{x} = 0.798$, $s_{intra} = 0.162$) and \vanilla{} ($\bar{x} = 0.787$, $s_{intra} = 0.152$) were the smallest yet very similar.

The analysis revealed a significant \kruwa{14.9504703937183}{4}{0.004805055} main effect of \dataset{} with a small effect size.
Post hoc tests confirmed significantly lower intra-prompt \rougel{} scores for \vanilla{} compared to both \json{} \mawiu{0.01156}{-0.0411}{-0.179}{} and \yaml{} \mawiu{0.03352}{-0.0431}{-0.162}{} with small effect sizes each.
\section{Discussion}\label{sec:discussion}
The controlled experiment conducted in the scope of this research investigated the effects of different prompt formats on the task performance, efficiency, and prompt stability in LLM coding tasks.
In this section, we will discuss the findings of said experiment, contrast them to our hypotheses, and derive implications for using LLMs in coding tasks.

\subsection{A Silver Lining in Task Performance}\label{sect:discussion/task_performance} 
Although we found slightly higher percentages of perfect \passrate{}s for the parsed levels of \dataset{} \json{}, \markdown{}, and \yaml{} compared to the baseline in \vanilla{}, our LLM-\tuned{} dataset led to the least performant code by far. On average, only \json{} and \markdown{} resulted in higher \passrate{}s than the baseline, none of which, however, were found to be statistically significant. Nevertheless, we did find that the choice of \dataset{} influenced the task performance (pass@1), as \tuned{} performed significantly worse than \json{}, \markdown{}, and even \vanilla{}---in the order of decreasing effect sizes---which is why we managed to reject $\text{H}_{1_0}$ and therefore accept $\text{H}_{1_\text{A}}$---although only partially in comparison to the baseline.

Because this slight gain in task performance for \json{} and \markdown{}, but not \yaml{}, coincides with their respective popularities and industry prevalences, we hypothesize that LLMs like ChatGPT can perform better when prompted in a more consistently formatted way it is familiar with based on its training data.
The \tuned{} dataset might have performed the worst because---although exhibiting increased internal consistency compared to the baseline---it was created using a second LLM by Mistral AI, which---in improving it based on its own training data and, thus, fitting it to it---might have inadvertently de-optimized it for ChatGPT. Furthermore, although HumanEval was initially designed not to be included in the training sets of code generation models, this might have changed in the past four years since its release in 2021---especially for ChatGPT, as both of them are by OpenAI.
Lastly, as we only verified the \tuned{} prompts on a sample basis, they might contain underlying inconsistencies we did not uncover.


Thus, we recommend at least trying out different prompt formats---primarily \json{}, for general-purpose LLMs like ChatGPT with non-specialized training data---as they might just improve the performance of LLM-generated code. However, before applying LLM tuning indiscriminately, both the modifying and receiving LLMs should be checked for practical alignment.

\subsection{Sky's the Limit for Code Generation Efficiency} 
In regard to processing efficiency, which we measured in the \genduration{}, we saw a significant decrease in the time required by the LLM to formulate its solutions to the given coding problems for all three reformatted levels of \dataset{} compared to the baseline, even despite their increased prompt lengths. Besides this, we found the same effect compared to the LLM-\tuned{} dataset, which, too, was on the smaller side. Notably, \yaml{}---possibly due to its compact size---reached the shortest average \genduration{} and biggest effect sizes in the post hoc tests.

This suggests that more rigorous structuring, like that of our reformatted prompts---without changing a single word from the descriptions---can aid LLMs in processing coding problems more efficiently, which leads us to partially accept H$_2$---only for the reformatted versions, not the tuned one, however. The prompt length might have been a secondary factor, allowing \yaml{} to prevail.



Based on these findings, we urge researchers to consider reformatting prompts in LLM coding scenarios in order to reduce the computational effort of their processing. Yet, care should be taken to ensure that the resulting prompt lengths do not balloon significantly, as this may counteract the benefits gained from the added structure.

\subsection{Code Efficiency's Still Up in the Air}
In terms of code efficiency, on average, both all (\evalduration{}) and all correct (\passduration{}) solutions derived from LLM-\tuned{} prompts were evaluated the fastest, because---although they had the lowest \passrate{}s---they had fewer extreme long runs (cf. \ival{/129} in Section~\ref{sect:results/EvalDuration}). The average \iv{Eval-} and \passduration{}s of the remaining levels of \dataset{} were similarly slower, with, however, no significant differences between their medians having been found. Thus, we fail to reject H$_3$'s null hypothesis regarding this second facet of efficiency.
Remarkably, the lower and upper bounds of \evalduration{}, when including only passing executions in the \passduration{} remained nearly unchanged.
Besides only a slight, albeit non-significant, uptick for \markdown{}, we did not uncover any---let alone significant---differences in \responselen{} either.

One possible explanation for \tuned{}'s arithmetic mean being so skewed compared to the very similar medians among all levels of \dataset{} might be that---percentage-wise---it failed more of the hardest, slowest-to-evaluate HumanEval coding problems, reducing its number of outlying \iv{Eval-} and \passduration{}s.


Thus, regarding the resulting code's conciseness and execution efficiency, we do not feel confident to derive best practices either way. However, these findings at least suggest that there is no wrong choice in this case.

\subsection{An Edge on Prompt Stability}
When it comes to prompt stability, we did uncover significant differences in the average of pairwise \rougel{} scores:
both \json{} and \yaml{} led to significantly higher \rougel{} scores than the baseline, with \markdown{} and---to a lesser extent---\tuned{}, too, exhibiting higher---although not significant---\rougel{} scores, and noticeably more perfect ones at that.
Therefore, H$_4$, too, can be accepted only partially for syntactic prompt stability.
Going beyond the LLM responses' character compositions, regarding semantic prompt stability, the results of \passrate{}---without significant differences, as described above in Section~\ref{sect:discussion/task_performance}---apply.

In these results, we see a confirmation of the expectable notion of more consistently structured prompts resulting in higher prompt stability. One important factor that might be at play is that the respective section headers of \json{}, \yaml{}, and \markdown{} act as an anchor by virtue of being always the same---especially compared to \vanilla{} and possibly \tuned{}, in some cases. The differences for \markdown{} may not have been significant due to its human-readable, syntactic-sugar-like formatting and representations for not-provided values, both of which \json{} and \yaml{} did not have.

We therefore highly recommend applying consistent formatting to prompts, as that reduces the variability of generated responses, in turn improving prompt stability. If not uniquely predefined otherwise by the usage scenario, going with the predominant \json{} format is a valid approach, as it resulted in the highest prompt stability in our experiment.
\section{Limitations \& Future Work}\label{sec:limitations_&_future_work}
We are convinced that the results presented in this paper provide valuable insights into the optimizability of prompts for LLM coding tasks. Yet, this work has limitations and further evinces avenues for future research.

\subsection{HumanEval \& the Levels of Dataset}
HumanEval is a synthetic dataset handwritten by OpenAI researchers for benchmarking LLMs, consisting of 164~relatively short, isolated coding problems. As this was the only dataset under investigation in our experiment, more research on how well our approach performs on coding problems from more complex datasets or real-life scenarios is advised. In particular, non-synthetic datasets would be of interest, for LLMs have been shown to perform better on synthetic ones under certain circumstances (cf.~\cite{SyntheticData2024}).

We used a model by Mistral AI for deriving the \tuned{} dataset to reduce resource costs. This might have resulted in reduced quality or at least adverse re-tailoring of the dataset for evaluation with ChatGPT. Future research should investigate different combinations of and model-intern LLM tuning to better understand the influences of LLM-specific fitting. It is not inconceivable that the training data of different LLMs has an effect on their alignment and agreement in real-world application scenarios.
In practice, conducting both the prompt tuning and processing of the coding problems with the same LLM---like ChatGPT---might yield the best results, or, possibly, if some LLMs turn out to be their perfect counterparts, they might achieve even greater performance in tandem.

Furthermore, our four derivatives of the HumanEval dataset were only verified on a sample basis, meaning that they might yet contain prompt-specific inconsistencies we overlooked---especially for the LLM-tuned one.
In general, all prompt optimization techniques discussed in this work---from LLM tuning to reformatting---could be integrated as a first, automatically occurring step into existing software development pipelines.

\subsection{Choice of LLM}
OpenAI's ChatGPT---particularily GPT-4o---is the most widely used LLM, which is why we decided to focus on it in the scope of this work in order to gain the most widely applicable and, thus, useful results.
However, as performance could vary based on the employed LLM, the guidelines derived from our findings should be verified using other LLMs to ensure generalizability for their real-world application, which would be particularly relevant for specialized coding LLMs.

\subsection{Correctness of Efficiency Measures}
As described in Section~\ref{sec:methodology/procedure}, the API requests were sent in sequence without counterbalancing, and, thus, the observed effects on the generation duration cannot be fully distinguished from possible temporal variations in API response times. Likewise, due to dependence on server traffic, the generation efficiency measure must not be interpreted as a universal truth, but only on a comparative time-local basis, as described in Section~\ref{sec:methodology/design/dvs}.\enlargethispage{1em} 

The evaluation of all LLM-generated responses was conducted on a local machine\footnote{The device the experiment ran on had the following specs:\begin{itemize}
    \item \textbf{CPU:} AMD Ryzen 7 7800X3D, 8C/16T, 4.20-5.00GHz
    \item \textbf{RAM:} Corsair Vengeance 64GB (2x32GB), DDR5-6000, CL30-36-36-76
    \item \textbf{GPU:} PNY GeForce RTX 4070 Ti XLR8 Gaming Verto Epic-X RGB Triple Fan, 12GB GDDR6X
\end{itemize}\vspace{-1em}} running Windows~11. To mitigate the introduction of noise for the time metrics, we refrained from using the machine for any other tasks whilst the experiment was running.

\subsection{Further Forays into Prompt Tuning \& Prompt Stability Evaluation}
Instead of using an LLM to tune the prompts directly, one could use a reinforcement learning approach like StablePrompt~\cite{kwon_stableprompt_2024} discussed in related work. Since said framework requires fine-tuning and---to our knowledge---there isn't a dataset fitted for our task, we propose using our automatic evaluation pipeline to fine-tune StablePrompt on a different dataset of a similar task. Our implementation of automatic task performance and efficiency evaluation could be used to generate the dataset needed for StablePrompt tuning. However, this would require the dataset of investigation to come with test cases like HumanEval does. At this point in time, our framework only works with HumanEval and datasets of the same format\footnote{Our implementation assumes the columns \texttt{task\_id}, \texttt{prompt}, \texttt{test}, and \texttt{entry\_point}, as present in the HumanEval dataset.}, with our pipeline needing to be adapted in other cases.

Considering the limited scope of this work, we focused on the classical measure of average pairwise ROUGE-L scores to quantify prompt stability as the likenesses of LLM-generated code responses over all 10~executions of a prompt.
Beyond this, future work should also investigate specialized measures like the Prompt Stability Score (PSS) in the context of LLM code generation, which can be calculated using the \href{https://pypi.org/project/promptstability/}{\texttt{promptstability} package}.
\section{Conclusion}\label{sec:conclusion}
In this work, we presented <<PromptResponse>>, an exploration of the effects of the structural make-up of prompts in LLM coding tasks in regard to task performance, generation and evaluation efficiency as well as prompt stability.
In a controlled experiment, we investigated these properties for five semantically identical versions of the HumanEval coding task dataset, including the original unaltered version, ones in the popular JSON, Markdown, and YAML formats, as well as an LLM-tuned one.
Our results indicate that reformatted prompts can lead to slight increases in task performance as well as more substantial ones in generation efficiency and syntactic prompt stability, with JSON standing out universally and Markdown and YAML in selected subcategories.
Regarding the LLM tuning of prompts, we see further research on the practical agreement of different LLMs in order, as our implementation only worsened the results.

\section*{Acknowledgements}
LanguageTool, QuillBot, and ChatGPT were used to spellcheck this paper before publication and to inform minor adjustments in phrasing and formatting.
We further ensured compliance with scientific notation using a developer build of <<Pre-Review>>, engineered at the Department of Chemistry at FU Berlin.

\section*{Supplementary Information}
The four derivative versions of HumanEval~\cite{HumanEval} as well as the code generation and evaluation pipeline developed for and employed in this work can be acquired from our \href{https://github.com/kuehnenr/PromptResponse/}{GitHub repository}.

\clearpage
\bibliographystyle{ACM-Reference-Format}
\bibliography{bibliography}

\end{document}